\documentclass[runningheads]{llncs}
\usepackage{graphicx}
\usepackage{amsmath,amssymb} 
\usepackage{color}
\usepackage[width=122mm,left=12mm,paperwidth=146mm,height=193mm,top=12mm,paperheight=217mm]{geometry}
\usepackage[utf8]{inputenc} 
\usepackage[T1]{fontenc} 
\usepackage{booktabs}
\begin{document}
\pagestyle{headings}
\mainmatter
\def\ECCV16SubNumber{***}  

\title{OmniCD: A Foundational Framework for Remote Sensing Image Change Detection Guided by Multimodal Semantics} 

\titlerunning{ECCV-16 submission ID \ECCV16SubNumber}

\authorrunning{ECCV-16 submission ID \ECCV16SubNumber}

\author{Chenhao Sun\thanks{Email: chenhaosunsgg@163.com}}
\institute{Wuhan University}

\maketitle

\begin{abstract}
Change detection (CD) in remote sensing is vital for applications such as urban monitoring and disaster assessment, yet traditional methods struggle with generalization across diverse scenarios. We present OmniCD, a foundational framework that unifies and enhances remote sensing CD through multimodal semantic guidance. OmniCD incorporates image and text prompts---such as textual descriptions, semantic maps, and geospatial metadata---into a unified architecture, supporting tasks from binary CD to zero-shot semantic change understanding. The framework integrates a hierarchical scene retrieval module and a change detection module, reinforced by a style disentanglement mechanism for improved cross-domain robustness. We further introduce RSITCD, a large-scale multimodal dataset with 300K+ annotated image-text pairs. Extensive experiments show that OmniCD achieves state-of-the-art performance across benchmarks, demonstrating strong adaptability and setting a solid foundation for general-purpose CD systems in remote sensing.

\end{abstract}

\section{Introduction}

Remote sensing technology, as a vital means of acquiring information about the Earth's surface, plays a crucial role in fields such as land resource management, environmental monitoring, disaster emergency response, urban planning, and agricultural assessment. Through satellites, UAVs, and aerial photography, it enables efficient and accurate observation of the Earth's physical, biological, and human features on a large scale. In recent decades, the rapid advancement of high-resolution satellite sensors, synthetic aperture radar (SAR), LiDAR, and multispectral/hyperspectral imaging has greatly enhanced the spatial, spectral, and temporal resolution of remote sensing data, offering unprecedented support for Earth observation.

With the continuous implementation of global Earth observation programs (e.g., NASA's Landsat series, ESA's Sentinel missions, and China's Gaofen series), the capability to acquire remote sensing data has been significantly improved. These datasets, characterized by high spatiotemporal resolution, multimodality, and multiscale information, have made it possible to dynamically monitor surface changes. Under the global context of climate change, rapid urbanization, and land use/land cover change (LUCC), how to efficiently extract valuable information from massive remote sensing data has become a core challenge in remote sensing science.

Change detection (CD), a key task in remote sensing image analysis, aims to automatically or semi-automatically identify changes in surface cover or target objects using multi-temporal imagery. It is widely applied in environmental monitoring, urban expansion analysis, agricultural yield estimation, and natural disaster assessment. For example, comparing pre- and post-disaster imagery allows rapid identification of affected areas, supporting emergency response and damage evaluation. Despite its importance, change detection faces challenges such as radiometric inconsistencies across images, diverse and multi-scale change types, and interference from pseudo changes due to environmental factors like shadows, atmospheric effects, and topography.

In recent years, deep learning has significantly advanced remote sensing interpretation~\cite{Kussul2017Deep,Li2022DeepSurvey}, achieving remarkable results in tasks like land cover classification, building extraction, object detection, and change detection. However, due to the variability in data characteristics across scenes, current models often struggle with generalization and require scene-specific designs with large annotated datasets, leading to high costs and limited scalability.

Against this backdrop, the emergence of large multimodal models (e.g., GPT-4~\cite{OpenAI2023GPT4}, BLIP~\cite{Li2022BLIP}, SAM~\cite{Alexander2023SAM}) provides a new paradigm for remote sensing change detection. These models feature flexible prompting mechanisms that allow users to guide task execution with simple instructions, greatly reducing dependence on annotated data and improving adaptability across scenarios. Inspired by this, this paper proposes a prompt-driven change detection method based on large multimodal models. By simply inputting multi-temporal images and a textual prompt specifying the desired change type, the model can efficiently identify changes across diverse environments with strong generalization and high accuracy, offering both theoretical and practical contributions to the development of intelligent remote sensing interpretation.The main contributions of this paper are as follows:

(1) We introduce the task of Open-Category Change Detection (OCCD) in remote sensing and construct a large-scale multimodal dataset, RSITCD, which covers diverse scenes and land-cover types, providing comprehensive and reliable support for OCCD research.

(2) We propose OminiCD, a multimodal semantic prompt-guided change detection framework tailored for OCCD. The framework adopts an end-to-end architecture and supports efficient and flexible change detection using either image-based or text-based prompts.

(3) Extensive experiments demonstrate that OminiCD consistently outperforms existing OCCD methods, and that RSITCD significantly enhances the performance of various models on open-category change detection tasks.

\section{Related Work}

\subsection{Remote Sensing Image Change Detection}

Remote sensing image change detection methods can be broadly categorized into traditional and deep learning-based approaches. Traditional methods include pixel-based and object-based techniques. Pixel-based methods focus on spectral differences between pixels to identify changes, using techniques like PCA~\cite{Deng2008PCA} and MAD~\cite{Marpu2011Improving}. While these methods improve feature extraction, they are susceptible to noise and radiometric variations, limiting their generalization ability. Object-based methods~\cite{Chen2012Object,Walter2004Object}, which segment images into objects for analysis, better handle noise and improve accuracy by leveraging spatial context. However, their performance depends on segmentation quality and faces challenges in large-scale data processing.

Deep learning-based methods have advanced the field significantly. Unlike traditional methods, deep learning can automatically extract complex features, making them more accurate and robust. CNN-based methods like UNet~\cite{Ronneberger2015UNet} and its variants, including STANet~\cite{Chen2020STANet} and SNUNet~\cite{Fang2021SNUNet}, use encoder-decoder architectures and attention mechanisms to enhance detection capabilities. Transformer models~\cite{Kolesnikov2021ViT,Chen2022ChangeTR} have also been incorporated, capitalizing on their ability to model long-range dependencies within the image. Combining CNN and Transformer networks~\cite{Liu2022CNNTransformer} has further improved change detection accuracy by capturing both local and global features.

Despite these advancements, deep learning-based methods still rely heavily on large labeled datasets, which are often difficult to obtain. To address this limitation, unsupervised and cross-domain methods have emerged, reducing the need for extensive labeled data and improving performance in varied conditions. However, these methods still struggle with data distribution shifts, which affect their transferability across different domains.

\subsection{Remote Sensing Foundation Model}

Recent advancements in natural language, visual, and multimodal base models have spurred their application in Earth observation, particularly in remote sensing. These models can be categorized into four types: visual base models, visual-language models, visual-geolocation models, and generative models.

Visual base models, inspired by computer vision, have evolved with the rise of contrastive learning and masked image modeling techniques, such as MAE~\cite{He2022MAE} and BEiT, to better analyze remote sensing data. Models like SeCo~\cite{Manas2021Seasonal} and CACo~\cite{Mall2023ChangeAware} capture land cover changes, while SpectralGPT~\cite{Hong2024SpectralGPT} introduces 3D masked image modeling for multispectral images. Models like SkySense~\cite{Guo2024SkySense} integrate spatiotemporal data, achieving breakthroughs in various tasks.

Visual-language models, extending from natural language processing, enhance image understanding through vision-language synergy. Models like RSGPT~\cite{Hu2023RSGPT}, RemoteCLIP~\cite{Liu2024RemoteCLIP}, and GeoChat~\cite{Kuckreja2024GeoChat} excel in tasks like cross-modal retrieval and object counting. Techniques like GRAFT~\cite{Mall2023RemoteAlign} reduce annotation costs by aligning geographic data with images for text-free image-text relationships.

Visual-geolocation models focus on geographic feature learning through contrastive learning between location and image encoders. Models like GeoCLIP~\cite{Cepeda2023GeoCLIP} and SatCLIP~\cite{Klemmer2023SatCLIP} improve image interpretation and support tasks like biodiversity classification and population density estimation.

Generative models, such as DiffusionSat~\cite{Khanna2024DiffusionSat}, address tasks like super-resolution and cloud removal. These models combine remote sensing images with geospatial data for improved image generation and restoration, showing great potential for future flexible solutions.

Despite progress, remote sensing models still lack a universal approach for change detection. Current models often fail to capture the required change information and multimodal data characteristics, highlighting the need for a specialized model with multimodal understanding for change detection.

\section{Methodology}

In recent years, the rapid advancement of multimodal foundation models has brought prompt-based visual task processing methods into the spotlight, offering new perspectives and potential for enhancing the flexibility and generalizability of change detection. However, most existing approaches remain limited to fixed categories or predefined labels, falling short in addressing real-world demands where change types are diverse and detection objectives require greater adaptability. Currently, there is a lack of end-to-end models specifically designed for prompt-guided remote sensing change detection.

To address this challenge, we propose a multimodal semantic prompt-guided change detection network based on a detector--guide collaborative architecture. This method allows users to input arbitrary semantic prompts to accurately identify changes of interest in remote sensing imagery.

As shown in Figure 1, the network consists of three main components: a feature extraction module, a Transformer-based guider module, and a pyramid scene parsing-based detector module. During inference, the model takes a pair of bi-temporal remote sensing images and a semantic prompt as input. These inputs are first processed by the feature extraction module to obtain image and prompt features. The image features are fed into both the guider and the detector, while the prompt features are passed into the guider only.

The guider uses the prompt features to refine the bi-temporal image features and generate an attention map highlighting the regions of interest. This attention map, along with the image features, is then passed to the detector. The detector identifies candidate change regions based on the temporal features and further refines them using the attention map to retain only the prompt-relevant changes, enabling prompt-guided change detection.

To further improve generalization, a style disentanglement module is introduced. It separates style features---related to imaging conditions such as sensor differences, illumination, or atmospheric effects---from the encoded image features, thereby suppressing pseudo-changes caused by such variations.

The following sections detail the design of each module in the proposed network.

\begin{figure}
    \centering
    \includegraphics[width=0.8\linewidth]{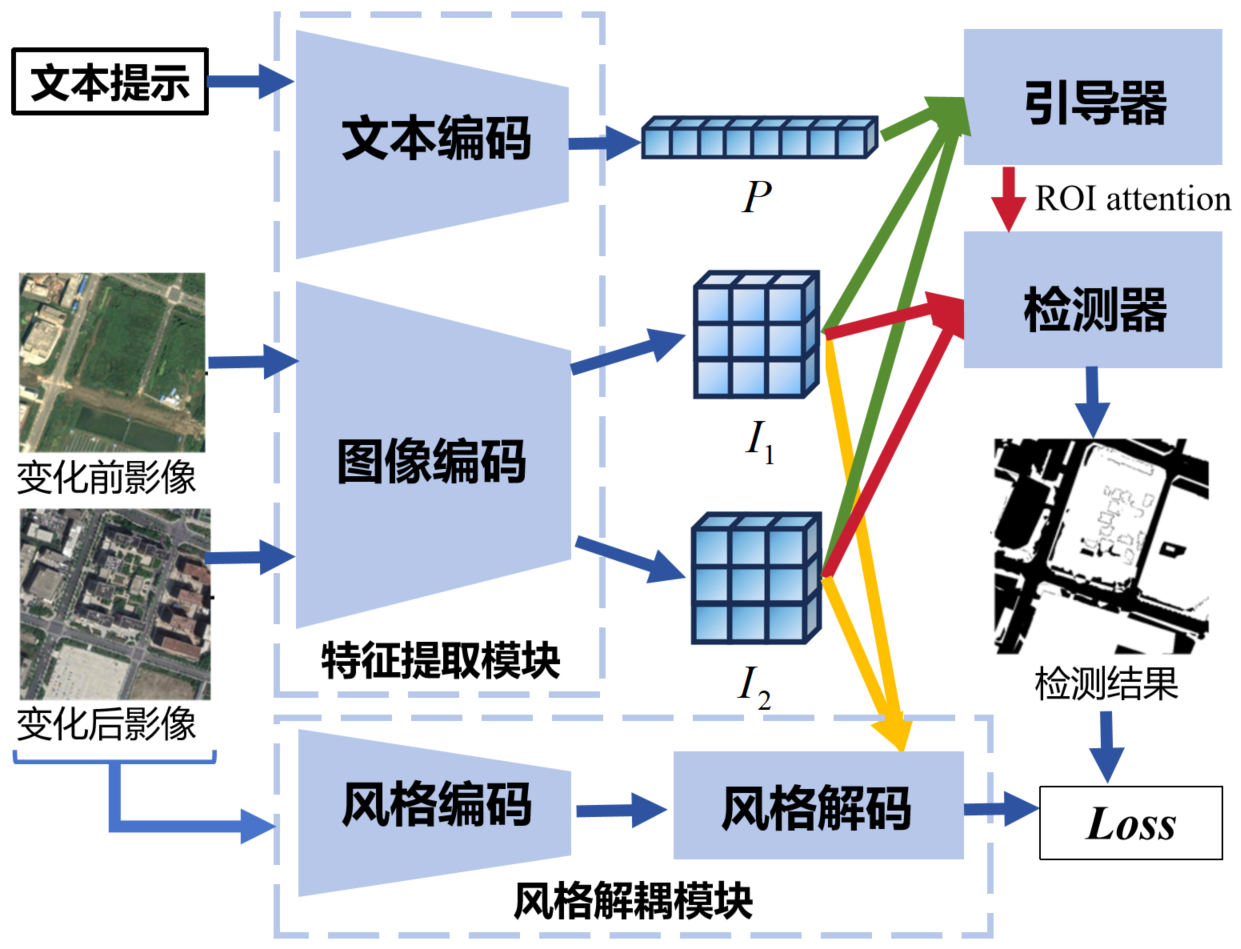}
    \caption{Framework diagram}
    \label{fig:enter-label}
\end{figure}

\subsection{Feature Extraction Module}
This module consists of two components: an image encoder and a text encoder.

For the image encoder, it can be any network that outputs a feature map of size $C \times H \times W$. To achieve scalability and fully leverage powerful pre-trained models, this work employs a Vision Transformer (ViT)~\cite{Kolesnikov2021ViT} pre-trained with MAE~\cite{He2022MAE}, with minimal adaptive modifications to handle high-resolution inputs. Specifically, ViT-H/16 is used, which contains $14 \times 14$ window attention and four evenly spaced global attention blocks. The output of the image encoder is the embedding of the input image after $8\times$ downsampling. The training image input resolution is $512 \times 512$, obtained by resizing and padding the shorter edge of the image; therefore, the image embedding has a size of $64 \times 64$. To reduce the channel dimension, a $1 \times 1$ convolution is applied to reduce the number of channels to 256, followed by a $3 \times 3$ convolution also with 256 channels. Each convolution operation is followed by layer normalization. Let the bitemporal images be $\mathbf{I}_1$ and $\mathbf{I}_2$, the image feature extraction process can be described by Eqs.~(1)--(2), where $C$ is the number of encoded channels, $\mathrm{LN}$ denotes layer normalization, $\mathbf{F}$ represents the ViT-encoded image features, and $\mathbf{E}$ denotes the final output image embeddings:

\begin{equation}
F_{\text{img}} = \text{ViT}(I) \in \mathbb{R}^{C \times 64 \times 64}
\end{equation}

\begin{equation}
I_{\text{emb}} = \text{LN}(\text{Conv}_{3\times3}(\text{Conv}_{1\times1}(F_{\text{img}}))) \in \mathbb{R}^{256\times64\times64}
\end{equation}

For the text encoder, the BERT~\cite{Devlin2018BERT} model is employed to generate text embeddings. BERT (Bidirectional Encoder Representations from Transformers) is a bidirectional self-attention model that is pre-trained using masked language modeling and next sentence prediction tasks, enabling it to understand the semantic meaning of words in context. In this work, the BERT-base model is used, which consists of 12 Transformer encoder layers, each with 12 self-attention heads, a hidden size of 768, and a total of approximately 110 million parameters. The text encoder first tokenizes the input text and encodes each token with word embeddings and positional embeddings, with a maximum sequence length set to 512 tokens. These embeddings are then processed through multiple layers of self-attention to capture contextual information in the text. Finally, the output is average-pooled to reduce the embedding dimension to 256. Let the input text prompt be $\mathbf{T}$ with tokenized length $L$, $\mathbf{F}_t$ denote the BERT-encoded features, and $\mathbf{E}_t$ denote the final output text prompt embedding. The text encoding process can be described by Eqs.~(3)--(4):

\begin{equation}
P_{\text{out}} = \text{AvgPool}(E_{\text{txt}}) \in \mathbb{R}^{N \times 256}
\end{equation}

\begin{equation}
I_{\text{emb}} = \text{LN}(\text{Conv}_{3\times3}(\text{Conv}_{1\times1}(F_{\text{img}}))) \in \mathbb{R}^{256\times64\times64}
\end{equation}

\subsection{Guide Module}

This module aims to efficiently map bitemporal remote sensing image embeddings and prompt embeddings to output masks (Region of Interest, ROI attention). This work refers to the Transformer-based segmentation model SAM~\cite{Alexander2023SAM} and modifies the standard Transformer decoder. To integrate prompt information with image embeddings, a learnable output token embedding is inserted among the prompt embeddings to guide the decoder output.

\begin{figure}
    \centering
    \includegraphics[width=0.7\linewidth]{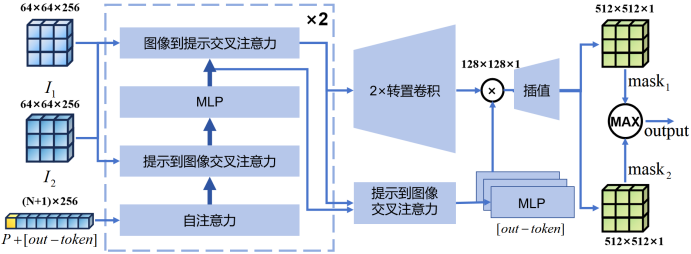}
    \caption{Guide Module}
    \label{fig:enter-label}
\end{figure}

The design of the decoder is shown in Figure~2. Let the input text prompt tokens be $\mathbf{T}$, and the image embeddings be $\mathbf{E}$, where $L$ denotes the tokenized text length. Each decoder layer performs the following four steps:  
\begin{enumerate}
    \item Self-attention on the prompt tokens.
    \item Cross-attention from prompt tokens (as queries) to image embeddings.
    \item Point-wise MLP updates for each prompt token.
    \item Cross-attention from image embeddings (as queries) to prompt tokens.
\end{enumerate}

Through these steps, the decoder effectively fuses prompt information with image embeddings and extracts masks for the regions of interest. In particular, during the cross-attention process, the image embeddings are treated as a set of $64 \times 64 \times 256$ vectors. Each self-attention, cross-attention, and MLP operation is equipped with residual connections, layer normalization, and a dropout of 0.1 during training. Each decoder layer outputs updated prompt tokens and image embeddings, which are passed to the next layer for further processing. In this work, two decoder layers are used. To ensure that the decoder fully utilizes geometric information, positional encodings are added to the image embeddings when they participate in attention layers.

After the decoder layers, the updated image embeddings are upsampled by $4\times$ through two $2 \times 2$ transposed convolution layers with stride 2 (resulting in 2$\times$ downsampling relative to the input image). Then, the upsampled image embeddings undergo attention with the prompt tokens again, and the updated output token embeddings are fed into a small three-layer MLP. This MLP outputs a vector matching the channel dimension of the upsampled image embeddings. By multiplying the MLP output with the upsampled image embeddings, a mask prediction for a single image is obtained, which is then interpolated to the original image resolution. Finally, the two sets of masks from the bitemporal images are combined using pixel-wise maximum to generate the final mask prediction as the ROI attention feature.

The Transformer in the decoder uses an embedding dimension of 256. The internal dimension of the MLP blocks is large (2048) but only applied to a small number of prompt tokens (usually no more than 20), so it does not impose a high computational burden. To further improve computational efficiency, in the cross-attention layers, the channel dimensions of queries, keys, and values are halved to 128. When the image embedding size is $64 \times 64$, this channel compression effectively reduces computation. All attention layers employ 8 attention heads to ensure stability and expressiveness in the multi-head attention mechanism. The transposed convolutions used for upsampling the output image embeddings use GELU activation and are separated by layer normalization after each convolutional layer.

\subsection{Detector Module}

The detector module is designed to efficiently identify all potential change regions in bi-temporal remote sensing images. By integrating the input guided attention feature map, it filters these regions and outputs change detection results that maintain the same resolution as the original images. The structure of the module is illustrated in Figure 3 and consists of the following main steps:

\begin{figure}
    \centering
    \includegraphics[width=0.75\linewidth]{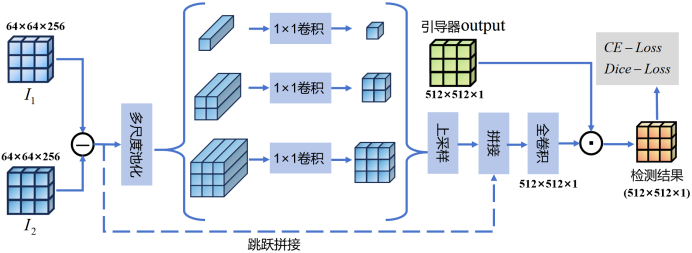}
    \caption{Detector Module}
    \label{fig:placeholder}
\end{figure}

First, the input bi-temporal images are encoded, and their features are subtracted channel-wise and converted to absolute values to obtain a change feature map. This step captures all the difference information between the two temporal images.

Next, the change features are fed into the Pyramid Scene Parsing (PSP) segmentation module for further processing. The PSP module employs a pyramid pooling strategy that performs multiple pooling operations at different scales (for example, 1 by 1, 2 by 2, 3 by 3, and 6 by 6). This operation divides the feature map into subregions of different sizes, enabling the integration of both global and local contextual information to address variations in object and scene scales. The pooled features are then refined through 1-by-1 convolutions for dimensionality reduction and subsequently upsampled to match the size of the original feature map. Afterward, these features are concatenated and fused with the original change feature map. The fused representation is further processed by convolutional layers, and a fully convolutional layer produces the final change detection results that align with the input image dimensions.

Finally, the resulting change detection map is multiplied pointwise with the guided attention feature map provided by the guidance module. This operation updates the change feature map based on the guided information, preserving only the change features in the regions of interest and effectively filtering for user-focused areas.

Through these steps, the detector module leverages multi-scale contextual information and guided attention to achieve fine-grained detection of change regions, producing high-precision change detection results.

\subsection{Style decoupling module}

The style decoupling module is designed to isolate style features that are related to imaging conditions from bi-temporal remote sensing images, thereby suppressing pseudo-changes caused by sensor discrepancies, illumination variations, or atmospheric disturbances. Its core idea is based on disentangled representation learning~\cite{Wang2024DisentangledSurvey}, guiding the model to decompose the input image into two orthogonal sub-features: content features and style features. Content features represent the semantic information of geographical entities (e.g., buildings, vegetation, roads) and are independent of imaging conditions, while style features capture dynamic factors in the imaging process (such as sensor parameters, seasonal lighting, and atmospheric scattering). Unlike content features that emphasize spatial details, style features focus on global understanding and remain spatially independent.

To achieve this, an additional pair of Siamese style branches is constructed on top of the feature extraction module, creating a style space for bi-temporal images that is independent of semantic content. Since content and style features are extracted separately from the input images, preserving their integrity is essential. A practical approach is to reconstruct the remote sensing image using its own content and style features. When the reconstructed image closely resembles the original input, it indicates that the extracted features sufficiently encode the image information. Therefore, an image reconstruction subtask is introduced into the style branch. This strategy enhances both content and style feature extraction by encouraging the recovery of the full image, thereby minimizing information loss.

\begin{figure}
    \centering
    \includegraphics[width=0.2\linewidth]{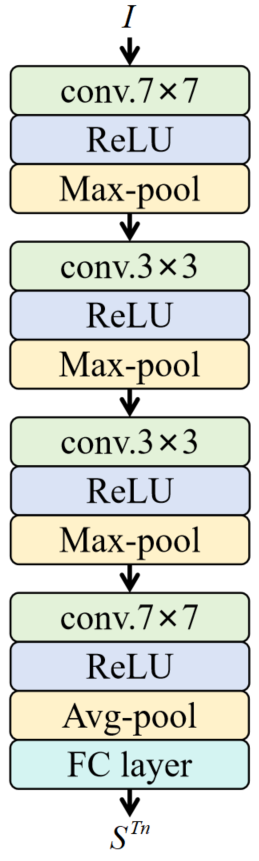}
    \caption{Encoder Structure of the Style Detection Module}
    \label{fig:placeholder}
\end{figure}

To ensure clear separation between content and style, two lightweight encoders are used to extract style features. Considering that the style encoders focus on spatially independent style representation, a series of convolutional blocks is employed to gradually reduce the spatial resolution of feature maps. As illustrated in Figure 4, each block consists of a convolutional layer with a kernel size of 3 or 7, followed by a ReLU activation and a pooling layer. Fully connected layers further compress the spatial dimensions to a single value, producing a style representation with the same dimensionality as the content features. This enables the encoder to focus on capturing global style information.

In addition, to restore the style of the image without altering its semantic content, Adaptive Instance Normalization (AdaIN)~\cite{Huang2017AdaIN}, a widely used technique in style transfer, is integrated into the reconstruction decoder. AdaIN transfers the style information into the content features by aligning their statistical properties.

\begin{figure}
    \centering
    \includegraphics[width=0.5\linewidth]{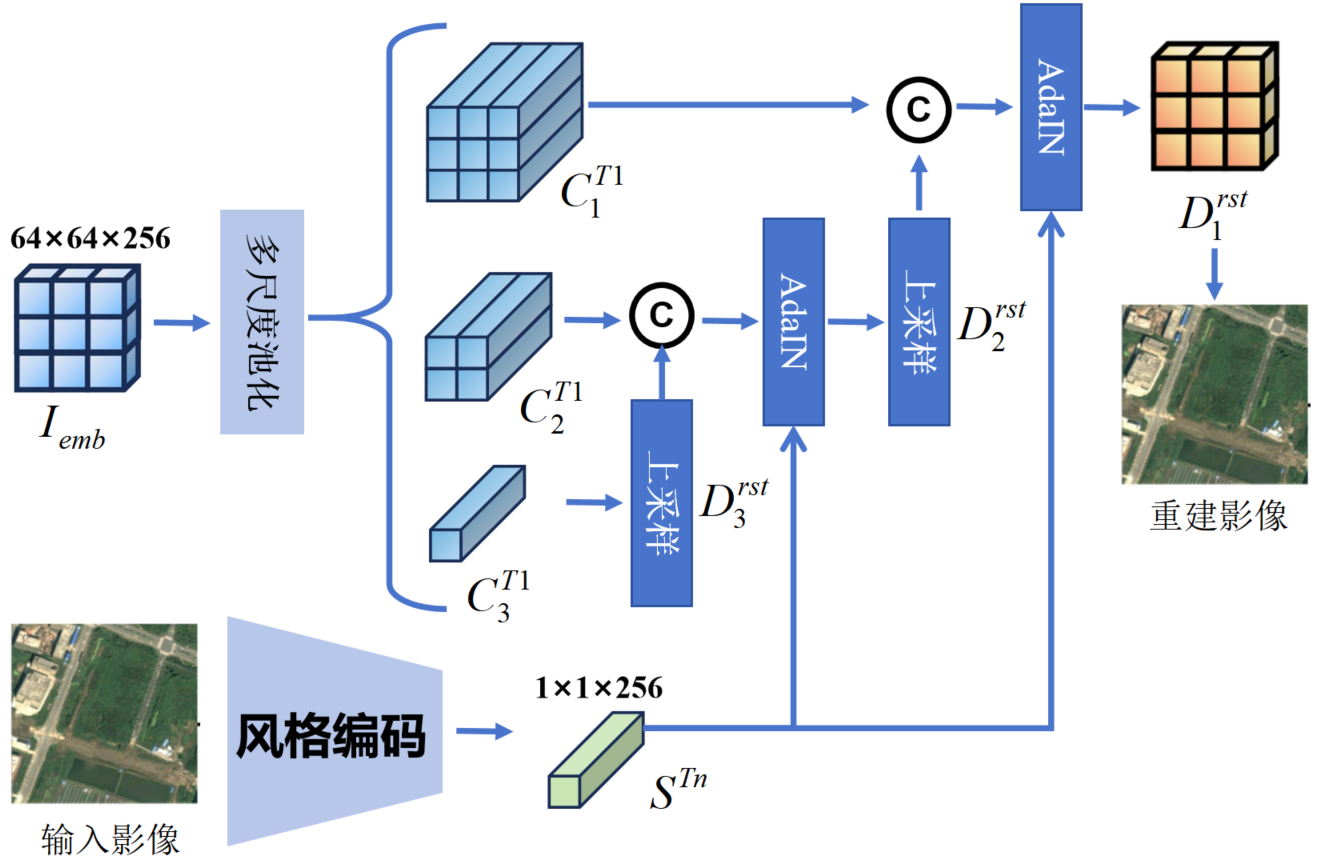}
    \caption{Decoder Structure of the Style Detection Module}
    \label{fig:placeholder}
\end{figure}

As shown in Figure 5, a multi-scale cascaded reconstruction strategy is adopted in the decoder. This process allows the decoder to fuse content and style features across multiple scales while progressively recovering the remote sensing image.

At each stage, the feature map from the previous decoding block is first upsampled and then concatenated with the content features. A convolution layer is used to fuse these features, and the fused result---together with the style features---is fed into AdaIN. The final output is activated by a ReLU function. This progressive decoding procedure, illustrated in Figure 6, enables the model to gradually integrate content and style information and reconstruct remote sensing images that maintain the same style as the original inputs.

\begin{figure}
    \centering
    \includegraphics[width=0.3\linewidth]{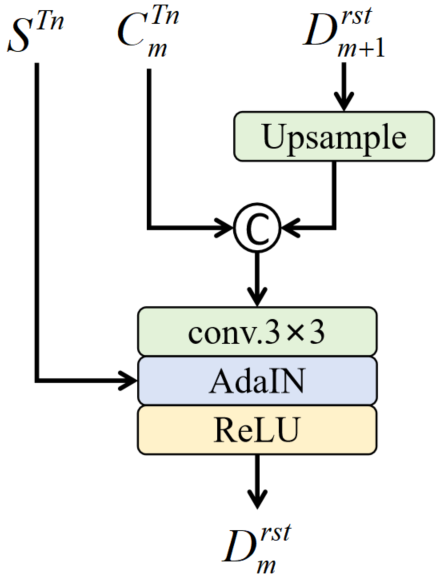}
    \caption{Decoding Block of the Style Detection Module}
    \label{fig:placeholder}
\end{figure}

\subsection{Loss Function}

To jointly optimize the change detection task and the image reconstruction task, this study introduces a multi-task loss function in which several loss components are combined and weighted by hyperparameters to balance their contributions. These components work together to guide the network toward learning both tasks in a cooperative manner.

The first component is the change detection loss. Considering that the primary goal is accurate binary change segmentation, a combination of Binary Cross-Entropy and Dice loss is employed to evaluate the correctness of the predicted change maps derived from bi-temporal images.

The second component is the feature separation loss. Since the framework uses two independent encoders to extract content and style features, this loss is introduced to strengthen the distinction between the two types of information. It encourages content and style features to remain linearly independent by penalizing their correlation.

The third component is the content similarity loss. Because bi-temporal images are captured from the same geographic location, the content features extracted from unchanged regions should remain consistent across time. This loss measures the similarity between content features by estimating the distributional distance between paired feature sets, using only the regions that exhibit no change.

The final component is the image reconstruction loss. To ensure that the extracted content and style features retain sufficient information from the input images, an L1-based reconstruction loss is applied. It guides the decoder to generate reconstructed remote sensing images that closely resemble the original inputs.

By integrating these losses into a unified multi-task objective, the model is encouraged to disentangle content and style effectively, achieve accurate change detection, and reconstruct images with high fidelity.

\section{Multimodal-Guided Change Detection}

In practical change detection scenarios, users can input arbitrary textual prompts without requiring additional annotated data, enabling the detection of changes related to specific semantic concepts. The detector--guidance architecture proposed in this work leverages prior knowledge learned from large-scale image--text datasets to interpret user-defined prompts and perform change detection based on bi-temporal images. During inference, as shown in Fig. 1, the textual prompt is encoded to generate guidance features, which interact with bi-temporal image features inside the guidance module to highlight regions relevant to the prompt's semantics and produce an attention-guided feature map. The attention features are then combined with image features and fed into the detector, which identifies initial change regions and filters them according to the semantic guidance, retaining only those changes aligned with the user's intent. This cross-domain text-guided mechanism gives the model strong generalization capability, allowing it to adapt to diverse scenarios without retraining for specific change types, thereby improving flexibility and applicability.

In addition to textual prompts, users may also be interested in more abstract or fine-grained visual concepts---such as specific architectural styles or object types absent from the training data---which can be difficult to describe precisely using natural language. Inspired by personalized modeling methods~\cite{Zhang2023PersonalizeSAM} that require no task-specific training, this work enables single-sample--based change detection for user-specified visual concepts, as illustrated in Fig. 7. Given a reference image and mask, the system generates a confidence map indicating the likely location of the target concept in the test bi-temporal images. Visual features are extracted from both the reference and test images using a pretrained encoder, and foreground features from the reference mask are used as local descriptors of the target concept. Cosine similarity between each descriptor and the test image features produces multiple confidence maps, which are normalized and aggregated into an overall confidence map representing the object's estimated location. This map is then fed into the guidance module to produce region-of-interest attention features, which guide the detector to obtain the final change detection results.
\begin{figure}
    \centering
    \includegraphics[width=0.8\linewidth]{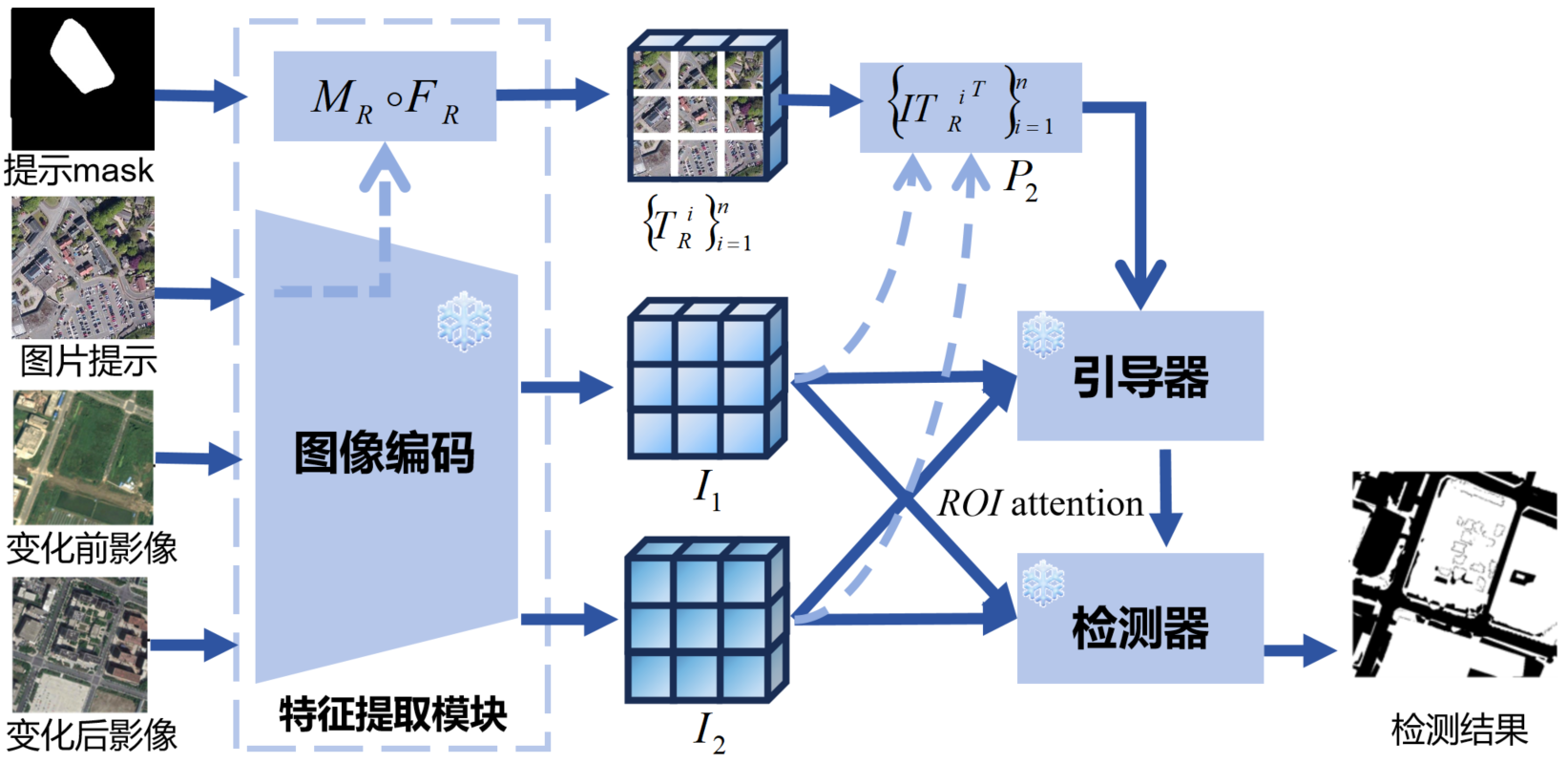}
    \caption{Illustration of Reference Image-Guided Method}
    \label{fig:placeholder}
\end{figure}

\section{RSITCD Dataset}

The proposed multimodal change detection model is a large-scale multimodal framework. Due to its substantial number of parameters, it requires a rich and diverse dataset to prevent overfitting and to enable effective cross-modal feature fusion, thereby enhancing the model's generalization, diversity, and robustness. However, existing change detection research lacks large-scale datasets suitable for training such models, and the available datasets vary significantly in organization, making them difficult to apply directly. To address this issue, we designed a unified data organization scheme specifically tailored for multimodal model training and integrated all publicly available change detection datasets, resulting in a large-scale remote sensing image--text multimodal change detection dataset (RSITCD).

The construction process of RSITCD is as follows:

First, we collected commonly used change detection datasets and selected those that contain bi-temporal remote sensing images, change region annotations, and semantic labels of the changed areas. Next, the original change labels were standardized into binary masks, which can be categorized into three types:

Single-type change datasets: These datasets contain only one type of change, represented as a binary mask. Examples include the building change detection datasets LEVIR-CD~\cite{Chen2020STANet} and WHU-CD. Pixel values typically take two values, e.g., {0, 1} or {0, 255}, which are unified into {0, 255}.

Multi-type changes with single-temporal labels: Each sample contains only one label map with multiple pixel values representing different change types, such as the CNAM-CD dataset. All possible combinations of change types are generated as binary masks to better distinguish different categories, enabling the model to learn richer and more complex feature representations.

Multi-type changes with bi-temporal labels: Similar to the previous type, but with two label maps corresponding to the two time points. Pixel-wise changes are extracted from these maps, and each type of change is converted into a binary mask for training. This allows the model to explicitly learn the transformation between different land cover classes.

After processing all change labels, corresponding textual prompts were generated for each sample. Considering the lack of domain-specific pretrained models, the training process requires high-quality and consistent data. Text prompts were produced using templates filled according to the binary labels. For example, the template ``Identify changes in \{change type\} in the image.'' was used. For single-type changes, the category is directly filled in; for multi-type changes, all combinations are mapped to the template; for bi-temporal multi-type changes, the text describes the transformation between classes, e.g., ``Identify changes in water bodies to bare land in the image.'' This approach ensures consistent formatting and structure, helping the model better understand the relationship between inputs and outputs, thus improving learning efficiency and prediction accuracy.

Finally, all bi-temporal images and corresponding labels were standardized to a resolution of 512*512. Images with higher resolution were cropped, while those with lower resolution were resampled using bilinear interpolation for images and nearest-neighbor interpolation for labels.

Through this integration process, the RSITCD dataset comprises 21 existing change detection datasets. Each sample includes a pair of bi-temporal remote sensing images, a change label, and a corresponding textual prompt, totaling 152,123 change detection samples (304,246 images) with semantic descriptions of changes.

\begin{figure}
    \centering
    \includegraphics[width=1\linewidth]{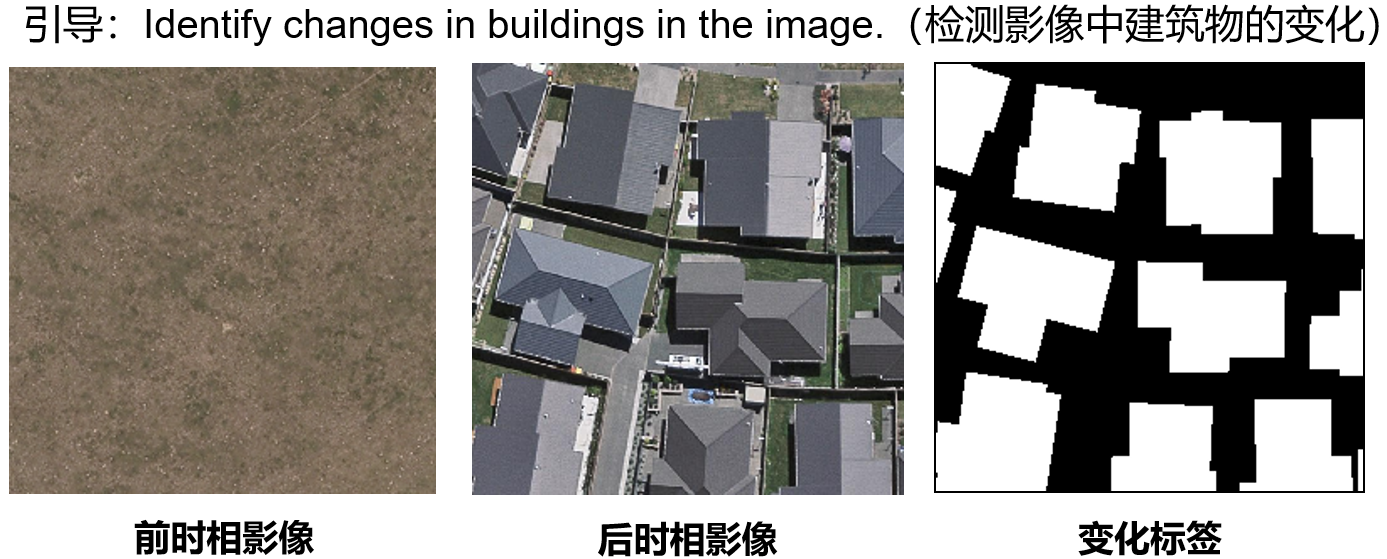}
    \caption{Sample Example from the RSITCD Dataset}
    \label{fig:placeholder}
\end{figure}

As shown in Fig. 9, the spatial resolution of RSITCD ranges from 0.03\,m to 2\,m, simulating the diversity of real-world applications. The dataset covers 17 land-cover categories and more than 40 types of changes, providing comprehensive support for model generalization, diversity, and robustness in practical change detection scenarios.

\begin{figure}
    \centering
    \includegraphics[width=1\linewidth]{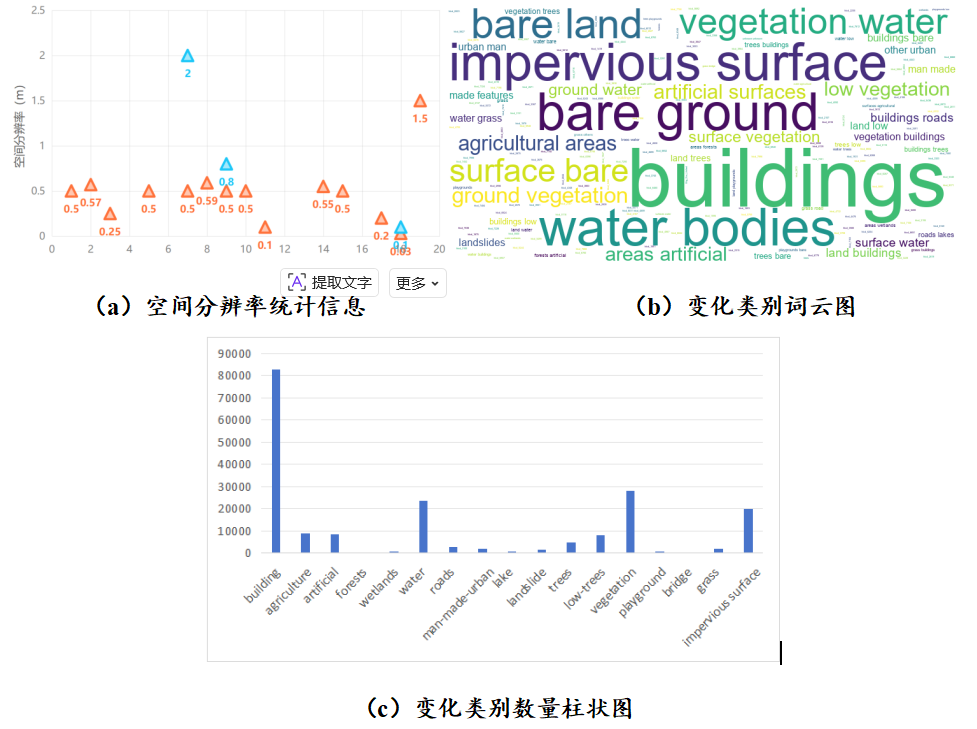}
    \caption{Statistical Summary of the RSITCD Dataset}
    \label{fig:placeholder}
\end{figure}

\section{Experiments}

\subsection{Datasets and Metrics}

\subsubsection{Datasets.}

The multimodal change detection test datasets used in this study consist of four datasets. Among them, two are building change detection benchmarks---LEVIR-CD~\cite{Chen2020STANet} and WHU-CD---and one is a landslide change detection benchmark, GVLM. The training sets of these three datasets are already included in the RSITCD dataset and are therefore used solely for evaluating the change detection capability of the proposed model. In addition, a self-constructed building change detection dataset, YRBCD, which is not included in the RSITCD dataset, is employed to further assess the model's generalization ability and cross-domain transfer performance.

The LEVIR-CD~\cite{Chen2020STANet} dataset is a large-scale building change detection dataset released by Beihang University in 2020. It covers 20 regions in several cities in Texas, USA, and contains 637 pairs of high-resolution (0.5\,m/pixel) Google Earth image patches of size 1024$\times$1024 pixels. The temporal interval between bi-temporal images ranges from 5 to 14 years. The dataset provides 31{,}333 annotated building change instances across various building types such as single-family houses, high-rise apartments, small garages, and large warehouses. In this study, the dataset is divided into training, validation and testing subsets using an 8:1:1 ratio. All images are cropped into 512$\times$512 patches through a sliding-window approach to meet the input requirements of the model.

The WHU-CD dataset, released by Wuhan University in 2018, records pre- and post-earthquake building changes in Christchurch, New Zealand. It consists of a pair of remote sensing images captured in 2012 and 2016, each with a resolution of 32{,}507$\times$15{,}354 pixels and a spatial resolution of 0.27\,m/pixel, covering approximately 20.5\,km$^2$. The dataset includes annotations for 12{,}796 building instances with detailed location and change information. In this study, the images are cropped into 256$\times$256 patches using a sliding-window strategy and randomly split into training, validation and testing sets according to an 8:1:1 ratio, containing 6{,}096, 762 and 762 pairs, respectively.

The GVLM dataset is a large-scale open-source landslide change detection dataset consisting of 17 pairs of very-high-resolution (0.59\,m/pixel) images obtained from Google Earth. The dataset spans 163.77\,km$^2$ and features landslide areas from different geographic regions, time periods, and surface conditions, exhibiting significant spectral heterogeneity. All images are cropped into patches of 512$\times$512 pixels and randomly divided into 1{,}398 training pairs, 175 validation pairs and 175 testing pairs.

Lastly, the YRBCD (Yangtze River Basin Change Detection) dataset is a self-constructed dataset developed for evaluating the cross-domain generalization performance of the proposed model. It reflects building changes across the Yangtze River Basin in China with an imaging interval of 5 years. The dataset is produced from GaoFen-6 (GF-6) satellite imagery that has undergone preprocessing and fusion, with a spatial resolution of 2\,m/pixel. Each image pair has a resolution of 512$\times$512 pixels. The dataset is randomly divided into 5{,}216 training pairs, 461 validation pairs and 461 testing pairs.
\begin{figure}
    \centering
    \includegraphics[width=0.8\linewidth]{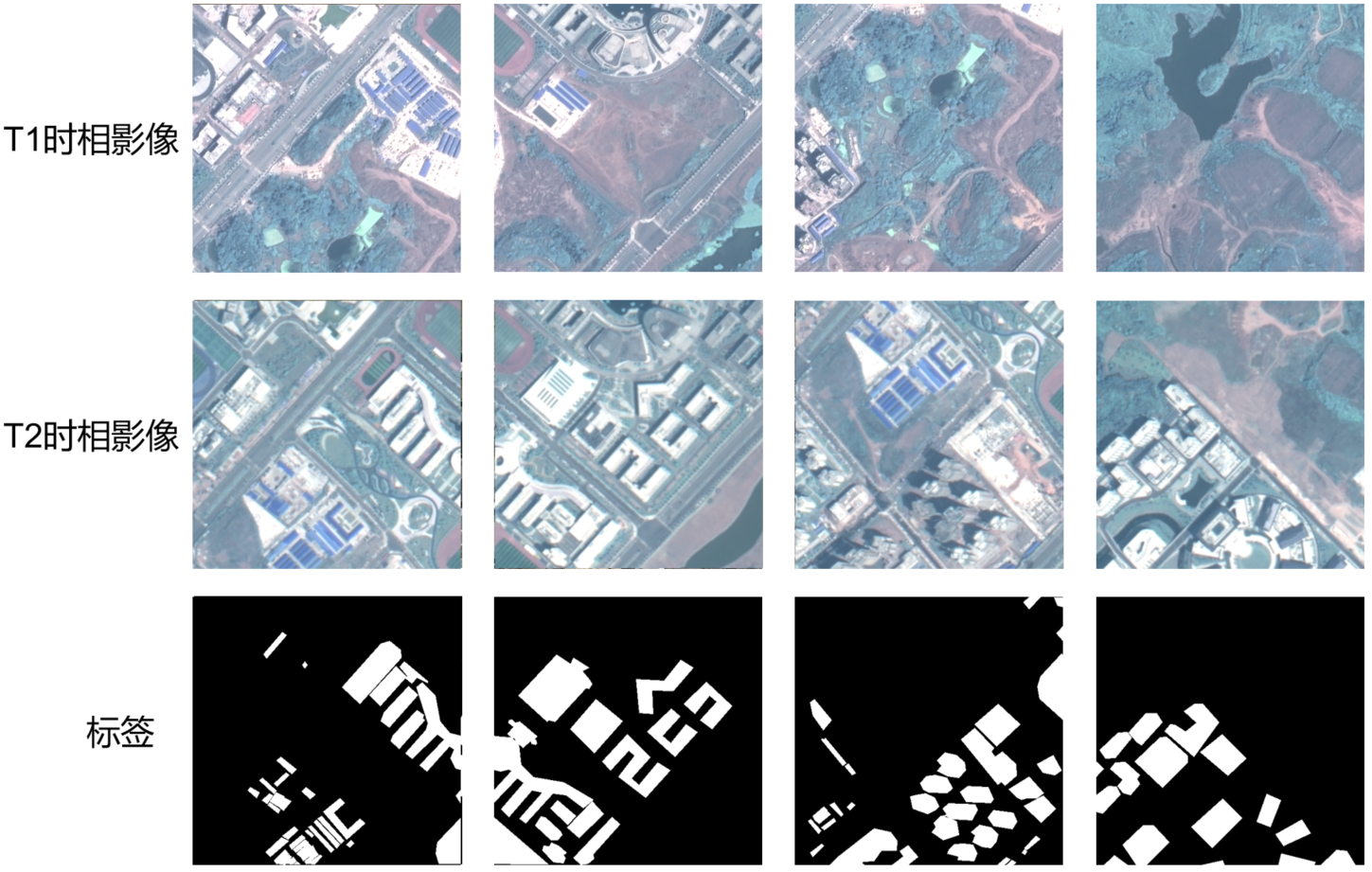}
    \caption{Sample images from the YRBCD dataset}
    \label{fig:placeholder}
\end{figure}

\subsubsection{Metrics.}

Remote sensing image change detection is essentially a pixel-level binary classification task. To comprehensively assess model performance, this study adopts five standard evaluation metrics: Precision, Recall, F1-score, Intersection over Union (IoU), and Overall Accuracy (ACC).

Precision measures how reliably the model identifies predicted change pixels, whereas Recall indicates how completely it captures the actual changed areas. The F1-score provides a balanced assessment by combining both Precision and Recall. IoU evaluates the spatial overlap between predicted and ground-truth change regions, reflecting the model's localization accuracy. ACC measures the overall correctness of pixel classification across both change and non-change categories.

Together, these metrics offer a robust and multi-perspective evaluation of the model's change detection capability.

\subsection{Implement Details}
All experiments in this section were conducted under a unified hardware environment. The computing platform is equipped with an Intel(R) Xeon(R) Gold~6278C (2.60\,GHz) CPU, two NVIDIA A40 GPUs, and 48\,GB of RAM. The implementation is based on the PyTorch deep learning framework, with PyTorch version~1.10.0 and torchvision version~0.11.0.

To directly compare the cross-domain retrieval capability of the model, all experiments in this section perform retrieval through inference using the pretrained model, without any additional training. During pretraining, the Adam optimizer was used with a learning rate of $1\times 10^{-4}$ and a batch size of 1 per GPU. The hyperparameters $\lambda_1$, $\lambda_2$, and $\lambda_3$ were all set to 0.1, and all other model hyperparameters and technical details remained consistent with the descriptions provided earlier.

\subsection{Results}
This section conducts experiments on two representative remote sensing change detection tasks---building change detection and landslide change detection---to evaluate the generalization capability of the proposed method across different scenarios and tasks. All experiments are performed by directly inferring with the pretrained model, without any additional training.

The comparison covers five representative change detection models:
(1) FC-EF~\cite{Daudt2018FCN}: an early-fusion fully convolutional encoder--decoder network that performs pixel-wise prediction in an end-to-end manner.
(2) FC-Siam-Conc~\cite{Daudt2018FCN}: a Siamese network with shared weights and feature concatenation to enhance cross-temporal contrast.
(3) FC-Siam-Diff~\cite{Daudt2018FCN}: a Siamese architecture that highlights change regions through feature differencing, reducing background interference.
(4) STANet~\cite{Chen2020STANet}: a spatiotemporal attention--based method designed to mitigate illumination variations and registration errors while strengthening temporal feature associations.
(5) SNUNet~\cite{Fang2021SNUNet}: a densely connected Siamese network with integrated channel attention, offering strong robustness in complex scenes.

The experimental results on the LEVIR-CD building change detection dataset are reported in Table 1. The prompt used in this experiment is ``Identify changes in buildings in the image.'' Since this section focuses on validating detection accuracy and visual performance under standard settings, the same prompt used during training is adopted here. The effect of varying prompts will be discussed in the following subsection.

\begin{table}
    \centering
    \caption{Comparison of Experimental Results on the LEVIR Change Detection Dataset (\%)}
    \label{tab:yrbcd_samples}
    \includegraphics[width=0.75\linewidth]{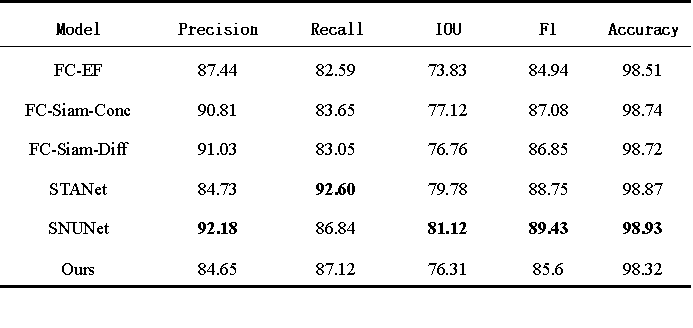}
\end{table}

The results in the table show that the proposed method achieves strong performance across key metrics---including Precision (84.65\%), Recall (87.12\%), IoU (76.31\%), F1-score (85.60\%), and Accuracy (98.3\%)---indicating that the model can effectively detect changes in remote sensing imagery. Compared with fully supervised baselines, the overall performance is largely comparable. The gaps from the best-performing supervised models remain within 8 percentage points (7.53 for Precision, 5.76 for Recall, 4.81 for IoU, 3.83 for F1, and 0.61 for Accuracy). These differences mainly arise because the proposed model is pretrained on a large-scale change detection dataset, enabling it to learn more generalizable change-sensitive features rather than dataset-specific patterns. As a result, achieving slightly lower performance than fully supervised models trained directly on LEVIR-CD is acceptable and expected. Moreover, variations in change appearance across datasets and differences in labeling practices also contribute to the remaining discrepancies.

\begin{figure}[htbp]
    \centering
    \includegraphics[width=1\linewidth]{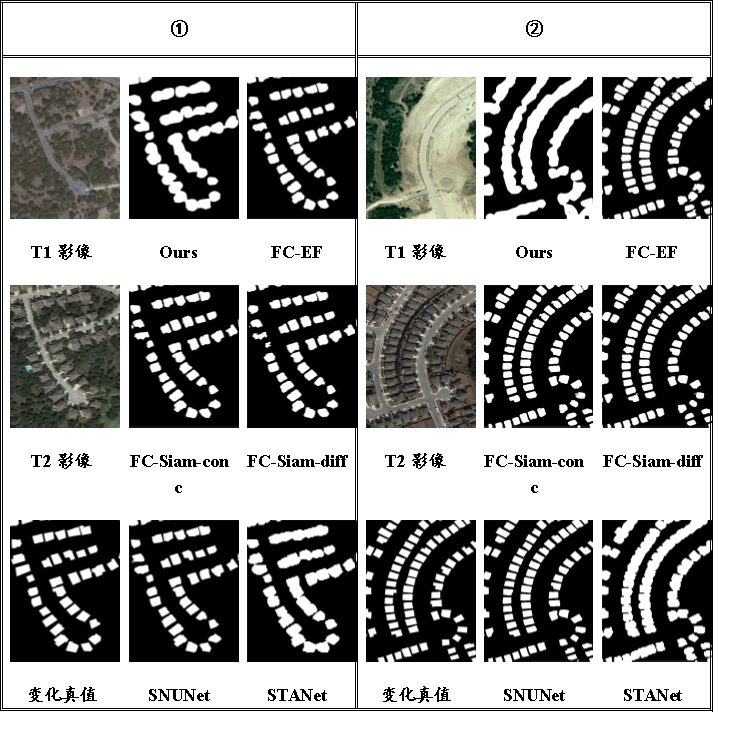}
    \caption{Comparison Results on the LEVIR-CD Dataset}
    \label{fig:placeholder}
\end{figure}

Figure 11 presents two representative examples from the LEVIR-CD dataset, illustrating typical building changes in dense urban areas. Compared with other methods, the change regions detected by our model appear smoother and less sharply delineated, although the overall localization accuracy remains high. This behavior is primarily attributed to the spatial pyramid structure embedded in the detector, which captures multi-scale information and integrates it through feature fusion. Such a design enhances robustness to variations in change size and morphology, enabling stable detection across diverse scenes, but may also result in softer boundaries compared with fully supervised models that overfit to dataset-specific textures and edge patterns. These smoother boundaries further reflect the model's strong generalization ability. Since it is pretrained on a large-scale change detection dataset, the model learns universal change representations rather than memorizing boundary characteristics unique to LEVIR-CD. Consequently, it maintains accurate localization across datasets with differing annotation styles, even if the boundary depiction is somewhat less crisp.

Table 2 reports the quantitative evaluation on the WHU-CD dataset using the prompt ``Identify changes in buildings in the image.'' The results show that our method outperforms FC-EF, FC-Siam-Conc, FC-Siam-Diff, and STANet in most metrics, and falls only slightly behind SNUNet, with performance gaps of 0.91\% in precision, 2.93\% in recall, 3.42\% in IoU, 1.92\% in F1-score, and 0.08\% in accuracy---all within 5\%. As noted earlier, these differences mainly arise from the fact that our model is pretrained on a large-scale change detection dataset, allowing it to learn more transferable and generalizable features rather than overfitting to the characteristics of WHU-CD. Although this leads to slightly weaker detection of fine details compared with fully supervised models trained exclusively on WHU-CD, our method still surpasses most of them, demonstrating strong capability in building change detection across domains.

\begin{table}
    \centering
    \caption{WHU Change Detection Dataset Experimental Comparison (\%)}
    \label{tab:yrbcd_samples}
    \includegraphics[width=0.75\linewidth]{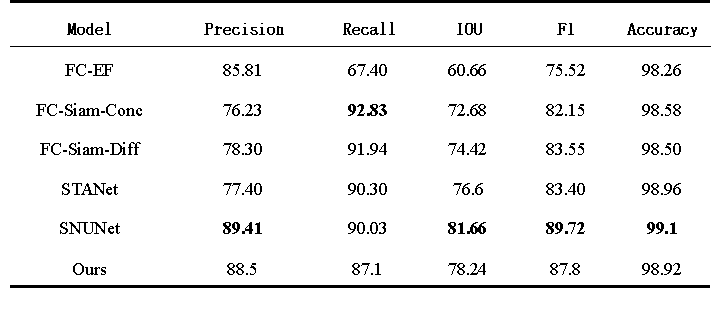}
\end{table}

\begin{figure}
    \centering
    \includegraphics[width=1\linewidth]{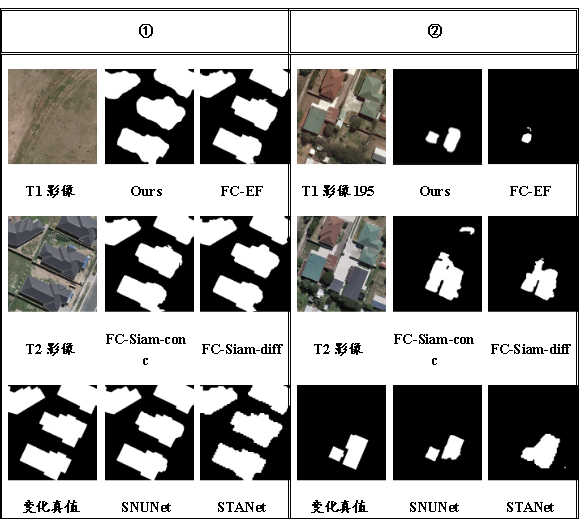}
    \caption{Comparison Results on the WHU-CD Dataset}
    \label{fig:placeholder}
\end{figure}

Figure~12 presents two representative samples from the WHU-CD dataset. Sample~1 illustrates a large newly constructed building area. Both supervised baselines and our method successfully detect the change regions. Consistent with the observations on the LEVIR-CD dataset, our method produces more complete regions with smoother boundaries, avoiding fragmentation and edge artifacts. In contrast, FC-EF and FC-Siam-Conc exhibit noticeable edge noise around buildings. Although our method is slightly less precise in boundary localization than supervised models, it effectively preserves structural integrity and delivers stable detection quality.

Sample~2 depicts a more challenging scenario involving two small changed buildings surrounded by vegetation-induced pseudo-changes and noticeable color differences between the two images. Only our method and SNUNet accurately detect both buildings. FC-EF misses the larger building and produces vegetation false alarms; FC-Siam-Conc and FC-Siam-Diff incorrectly classify paved ground and vegetation as changes; STANet roughly captures the building region but suffers from boundary blur and ring-shaped false detections; SNUNet performs well but still misses part of the right building. In contrast, our method precisely localizes the new constructions while suppressing pseudo-changes, demonstrating strong robustness in complex environments.

Table~3 reports the quantitative results on the GVLM landslide dataset, used to evaluate the model's multi-scenario capability. With the prompt ``Identify changes in landslide in the image.'', our method outperforms FC-EF, FC-Siam-Conc, FC-Siam-Diff, and STANet across most metrics, and is only slightly inferior to SNUNet (precision $-2.44\%$, recall $-1.95\%$, IoU $-2.01\%$, F1 $-2.15\%$, accuracy $-1.15\%$), with all differences within $5\%$. These results confirm the strong generalization ability of our method for change detection across diverse tasks and scenes.

\begin{table}
    \centering
    \caption{Comparison on the GVLM Change Detection Dataset (\%)}
    \label{tab:yrbcd_samples}
    \includegraphics[width=0.75\linewidth]{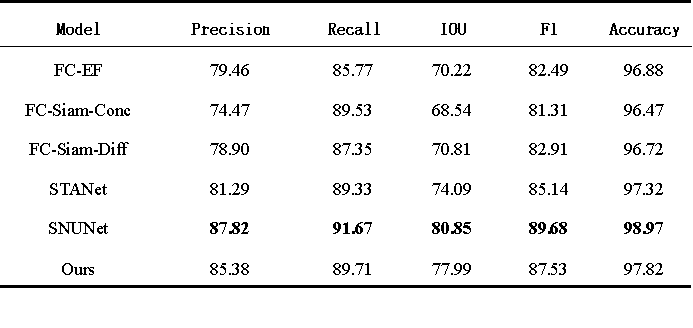}
\end{table}

\begin{figure}
    \centering
    \includegraphics[width=1\linewidth]{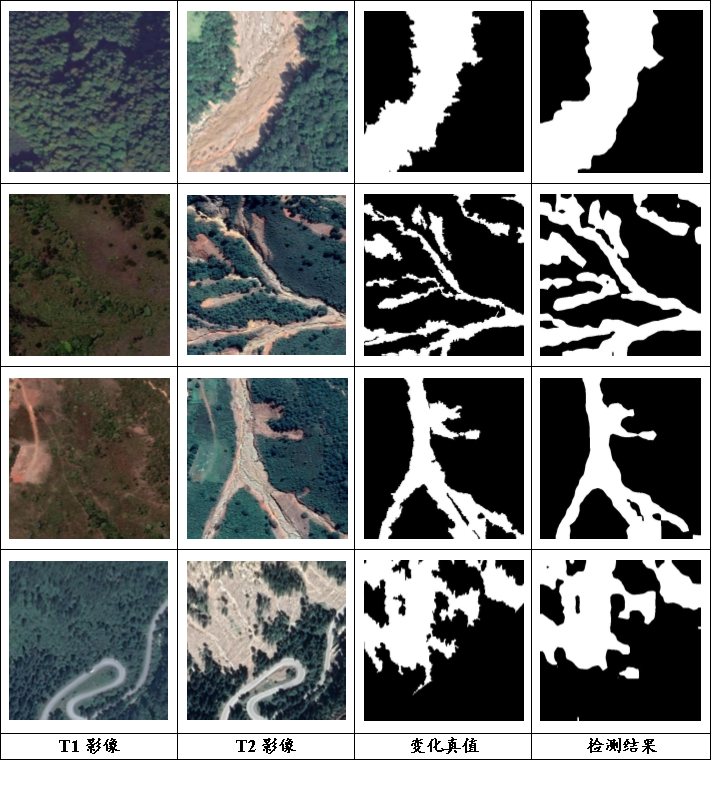}
    \caption{Detection Results on the GVLM Dataset}
    \label{fig:placeholder}
\end{figure}

To further assess the model's generalization and plug-and-play capability beyond the datasets used during pretraining, a cross-domain transfer experiment was conducted. Although the proposed approach performs well on LEVIR-CD, WHU-CD, and GVLM, these datasets are partially included in the RSITCD pretraining corpus, making it difficult to fully evaluate how the model behaves in truly unseen environments. Therefore, we select the real-world YRBCD dataset---completely excluded from training---to examine the model's zero-shot performance when directly applied to new domains.

To evaluate cross-domain robustness, two state-of-the-art unsupervised building change detection models are included for comparison, along with the supervised FC-EF model as a baseline. FC-EF is tested under two settings---trained on YRBCD and trained on LEVIR-CD---which helps quantify both the difficulty of the target dataset and the adaptability of traditional supervised methods under domain shift. The SCM model~\cite{Tan2024SCM} leverages SAM and CLIP to address multi-scale and semantic inconsistencies, while AnyChange~\cite{Zheng2024SegmentAnyChange} supports zero-shot change detection and provides both instance-level and pixel-level masks through automatic, semi-automatic, and interactive modes. Together, these baselines offer a comprehensive foundation for evaluating the cross-domain capability of the proposed method.

The YRBCD dataset presents substantial challenges. Its positive-to-negative sample ratio exceeds 1:50, reflecting an extreme imbalance that is closer to real operational settings. In addition, due to sensor variation, its visual style differs significantly from standard benchmarks, making feature transfer difficult. Clouds and shadows introduce further occlusion and pseudo-change noise, complicating boundary identification and semantic consistency. These properties make YRBCD a highly demanding benchmark for testing generalization and robustness.

For evaluating the proposed model, two types of prompts are employed: a text-based prompt and a reference-image prompt. For text prompting, we use ``Identify changes in buildings in the image.'' For reference prompting, a sample image from the test set is selected, and masks are manually drawn over several buildings to indicate the target class. This setup allows us to systematically examine how different modalities of prompts influence zero-shot change detection performance.

\begin{figure}
    \centering
    \includegraphics[width=0.5\linewidth]{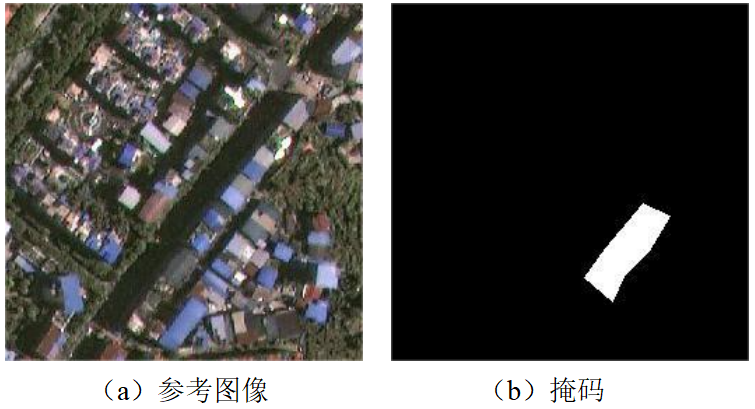}
    \caption{Reference-image prompting}
    \label{fig:placeholder}
\end{figure}

Table 4 presents the quantitative evaluation results on the YRBCD dataset. The proposed model is tested using two types of prompts---text prompts (T) and reference-image prompts (P). As shown in the table, our model substantially outperforms existing unsupervised methods under both prompting strategies. Without any domain-specific training, it achieves approximately 60\% precision and recall, around 40\% IoU, and over 99\% accuracy. These results indicate that the model can reliably identify and localize building changes while effectively suppressing false positives and adapting to significant variations in imaging style, demonstrating strong plug-and-play capability.

Although the performance of our method is still lower than that of fully supervised models trained directly on the YRBCD dataset---with a gap of roughly ten percentage points---this difference remains acceptable given the cross-domain setting and practical deployment considerations. Furthermore, the table shows that a supervised model trained solely on the LEVIR-CD dataset performs extremely poorly on YRBCD: its precision, recall, and IoU are all very low, indicating an inability to detect meaningful changes. Its accuracy remains above 90\%, but this is largely due to the severe class imbalance (positive samples $<$1:50), where predicting non-change regions dominates and masks its failure to detect true changes.

These findings highlight the limited cross-domain generalization capability of traditional supervised models. In contrast, the proposed method demonstrates markedly stronger robustness and generalization when transferred to entirely new datasets and imaging conditions.

\begin{table}
    \centering
    \caption{Comparative Experiments on the YRBCD Dataset}
    \label{tab:yrbcd_samples}
    \includegraphics[width=0.75\linewidth]{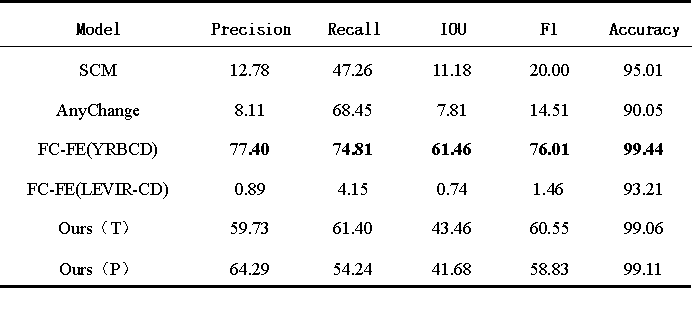}
\end{table}

To further demonstrate the effectiveness of reference-image guidance, two abstract building categories were selected for reference-based change detection. Figure~15 presents two representative examples from the YRBCD dataset. In Sample~1, the reference image corresponds to a stadium. The results show that, compared with text prompts, reference images enable more precise identification of the target building type (e.g., stadiums) while effectively filtering out irrelevant building changes. This confirms the advantage of reference-image guidance in fine-grained category detection and highlights its potential to broaden the scope of change-detection tasks.

Similarly, Sample~2 uses a circular building as the reference. The model successfully captures the unique structural characteristics of circular buildings---such as the closed ring shape---and accurately detects corresponding instances in the dataset. Together, the two examples illustrate that reference-image guidance adapts well to different building morphologies and supports more flexible and open-ended change-detection categories.

\begin{figure}
    \centering
    \includegraphics[width=1\linewidth]{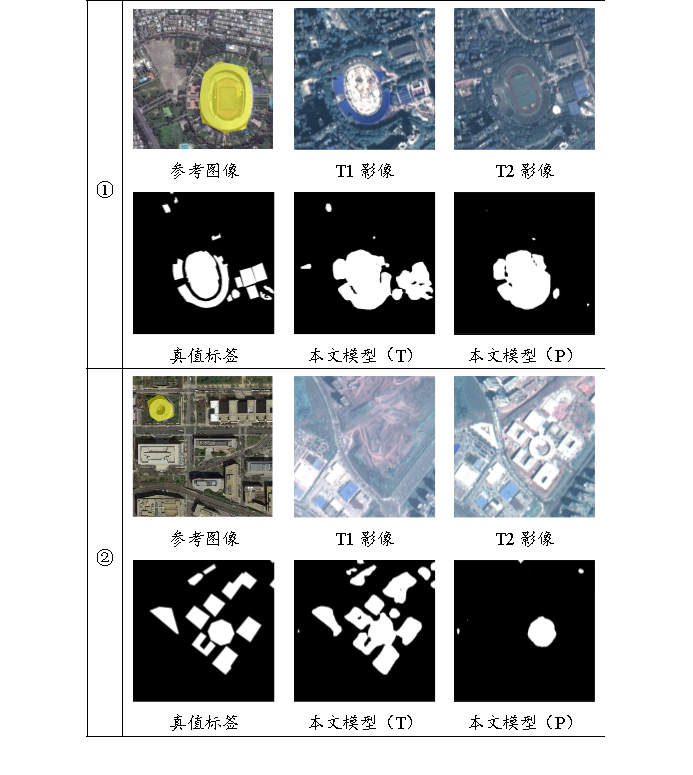}
    \caption{Detection Results on YRBCD with Reference Images}
    \label{fig:placeholder}
\end{figure}

\section{Conclusion}

This work focuses on advancing remote sensing image change detection by integrating state-of-the-art multimodal vision--language techniques. We propose a highly accurate, generalizable, and efficient change detection framework designed for strong adaptability across diverse scenarios and plug-and-play deployment. The main contributions are summarized as follows.

First, we develop a multimodal prompt-guided change detection framework that integrates multimodal prompting with hierarchical change analysis. By supporting both image-based and text-based prompts, the framework enables precise change localization and significantly enhances interactivity and adaptability across varied application contexts.

Second, we construct a large-scale multimodal remote sensing change detection dataset, RSITCD, containing more than 300,000 high-resolution images spanning over ten land-cover categories. The dataset provides abundant heterogeneous change information, enabling the model to learn robust and generalizable representations for complex real-world scenarios.

Third, we propose a prompt-driven multimodal change detection network based on a guider--detector architecture. By leveraging either reference images or textual descriptions as guidance, the method supports diverse change detection tasks and can be easily adapted to different environments, offering an efficient and flexible solution for large-scale deployment.

Extensive experiments demonstrate that the proposed framework consistently achieves strong performance in terms of accuracy, robustness, and generalization, outperforming existing unsupervised methods and providing new insights into the application of large multimodal models for remote sensing change detection.

Looking ahead, several directions merit further exploration. First, although the proposed model exhibits promising cross-domain transferability, future work should evaluate its performance on a broader range of datasets with varying resolutions, sensors, and scene distributions, potentially incorporating adaptive learning strategies to further improve transfer performance. Second, since the current architecture prioritizes generalization and model capacity, computational efficiency remains an area for improvement; future research may focus on designing more compact and efficient architectures for deployment in resource-constrained environments. Third, emerging AI techniques---such as state-space models and diffusion models---offer new opportunities for enhancing multimodal change detection and exploring generative representations of change patterns. Finally, expanding and openly releasing the RSITCD dataset while improving category balance and scene diversity will support community collaboration and foster the continued development of multimodal remote sensing change detection technologies~\cite{Huang2014Novel}.

\clearpage
\bibliographystyle{splncs}
\bibliography{egbib}
\end{document}